\documentclass{article} % For LaTeX2e
\usepackage{preprint,times}

\usepackage{amsmath,amsfonts,bm}

\def\eqref#1{equation~\ref{#1}}
\def\1{\bm{1}}

\DeclareMathAlphabet{\mathsfit}{\encodingdefault}{\sfdefault}{m}{sl}
\SetMathAlphabet{\mathsfit}{bold}{\encodingdefault}{\sfdefault}{bx}{n}

\usepackage{hyperref}
\usepackage{url}
\usepackage{comment}
\usepackage{makecell}
\usepackage{siunitx}
\usepackage{xfp}      % for \fpeval (floating-point math)
\usepackage{array}
\usepackage{float}
\usepackage{verbatim}
\usepackage{subcaption}
\usepackage{mathtools}
\usepackage[table]{xcolor}
\usepackage{amssymb}
\usepackage{multirow}
\newcommand{\dlt}[1]{\,{\scriptsize$(#1)$}}

\usepackage{cleveref}
\usepackage{booktabs}
\usepackage{verbatim}
\usepackage{wrapfig}

\title{\textbf{Out-of-Distribution Object Detection in Street Scenes via Synthetic Outlier Exposure\\ and Transfer Learning}
}
\author{
Sadia Ilyas$^{1,2}$\quad Annika M\"utze$^{1}$\quad Klaus Friedrichs$^{2}$\quad
Thomas Kurbiel$^{2}$\quad Matthias Rottmann$^{1}$\\[0.8ex]
$^{1}$Osnabr\"uck University\qquad
$^{2}$Aptiv Services Deutschland GmbH
}

\begin{document}

\maketitle

\begin{abstract}
Out-of-distribution (OOD) objects are often silently overlooked by object detectors, which produce no detection for them, leaving downstream modules with nothing to act on and no indication that anything was missed. 
A reliable object detector must detect OOD objects by localizing and correctly classifying them as OOD.
Existing approaches primarily focus on OOD detection for image classification or rely on auxiliary branches, and typically do not treat in-distribution (ID) and OOD objects in a unified manner. 
We present \textbf{SynOE-OD}, a \textbf{Syn}thetic \textbf{O}utlier-\textbf{E}xposure-based \textbf{O}bject \textbf{D}etection method,
which uses text-guided inpainting and an open-vocabulary object detector to generate and verify synthetic outliers offline. These are then used to fine-tune an object detector of choice with an added OOD class, which requires no auxiliary branch at inference time. 
On the OoDIS benchmark, SynOE-OD outperforms the strongest open-vocabulary baseline, reaching AP$_{50..95}$ of $57.6$\,\%, $29.4$\,\% and $39.2$\,\% on RoadObstacle, RoadAnomaly and Fishyscapes, against $48.5$, $22.7$ and $22.7$. This state-of-the-art performance holds for both fine-tuned closed-set and open-vocabulary detectors while maintaining competitive ID performance in street scenes.
\end{abstract}

% problem statement -> gap -> solution -> results

\section{Introduction}
\label{sec:intro}

Classical deep object detectors are typically trained on a fixed set of classes, 
and objects that appear outside this label space are assigned to the background~\cite{dhamija2020overlooked}. 
Hence, object detectors explicitly learn not to detect unlabeled object classes, in particular classes that are semantically out-of-distribution (OOD).
In safety-critical applications, e.g., autonomous driving~\cite{henriksson2023out} or 
medicine~\cite{gutbrod2025openmibood}, an object that is never localized leaves 
%no room for 
the downstream system nothing to act on. Detecting OOD objects matters as much as 
detecting in-distribution (ID) objects \cite{bogdoll2022anomaly, shoeb2025out}.

A significant amount of work addresses OOD detection for image classification \cite{hendrycks2016baseline, liang2017enhancing, lee2018simple, sun2022out, ren2019likelihood, liu2020energy, oberdiek2022uqgan} 
and semantic segmentation \cite{chan2021segmentmeifyoucan, zhao2024segment, rai2023unmasking, blum2019fishyscapes, bevandic2019simultaneous},
while OOD object detection remains comparatively understudied~\cite{du2022vos, ilyas2024potential}.
Most methods for OOD object detection~\cite{du2022vos, du2022siren, wilson2023safe} train  
a detector on ID data and add an auxiliary branch that scores each predicted 
box as ID or OOD. Such methods then inherit the same blind spot of classical 
object detectors, since the score only applies to boxes the detector predicts. 
OOD objects are thus recognized only if the detector happens to 
localize them, and their localization is neither learned nor evaluated. 
OoDIS~\cite{nekrasov2025oodis} evaluates the detection of anomalies at the instance level 
in street scenes, which makes it possible to measure OOD object detection performance.

\begin{figure}[t]
  \centering
\includegraphics[width=\linewidth]{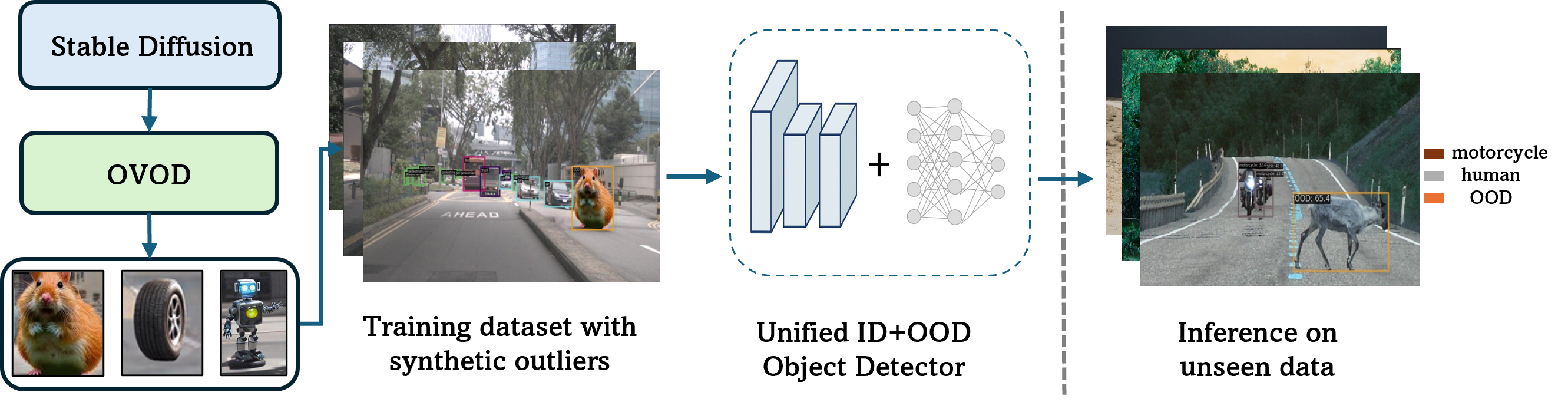}

   \caption{Overview of the SynOE-OD method, where an object detector is fine-tuned to jointly detect ID and OOD objects. NuImages is the training set augmented with synthetic outliers. At inference time, the detector is evaluated on an unseen real-world test set, e.g., RoadAnomaly. 
   }
   \label{fig:main_t1}
\end{figure}

Outlier Exposure (OE)~\cite{hendrycks2018deep} exposes a model to auxiliary outlier data
so that OOD inputs receive high uncertainty scores. VOS~\cite{du2022vos} adapts this idea to detection by synthesizing virtual outliers in feature space. 
Its evaluation follows the Pascal VOC/COCO setting~\cite{everingham2010pascal, lin2014microsoft}, where ID and OOD objects do not coexist in the same image. 
Recent advances in image generation via Stable Diffusion~\cite{rombach2022high, runwayml2023inpainting} make OE more scalable, 
as outliers can now be directly inserted into ID scenes. 

We therefore present SynOE-OD, illustrated in \cref{fig:main_t1}, which incorporates synthetic 
outliers directly into the image space and fine-tunes object detectors to jointly 
detect ID and OOD objects. 
Unlike methods that score the ID detector's predicted boxes as 
ID or OOD, SynOE-OD supervises outliers as detection targets in the same manner as any ID class, 
and thereby learns to detect OOD objects that would otherwise be assigned to the background.
Specifically, Stable Diffusion inpaints outliers at plausible locations 
in street scenes by replacing ID objects or filling free road regions,  
while Grounding DINO (GDINO)~\cite{liu2024grounding} annotates them with 
precise boxes and accurate labels. An object detector of choice 
is then fine-tuned via transfer learning with an additional $(n{+}1)$-st OOD class, 
where $n$ denotes the number of ID classes.
This formulation matches the requirements of applications such as autonomous driving, where known 
and unknown objects must be localized and classified in the same scene. 
In practice, OE is known to be sensitive to the choice of outlier data~\cite{wang2023dal, zhang2023mixture}; 
its performance degrades when the training outliers do not resemble the test-time anomalies. 
We study this dependence for synthetic outliers in the object detection setting for street scenes.

\noindent
Our key contributions are as follows: 
\begin{itemize}
    \item We introduce SynOE-OD, which extends closed-set and open-vocabulary object detectors to jointly detect
    ID and OOD objects, requiring neither real OOD data nor an auxiliary branch at inference time. 
    \item SynOE-OD sets a new state-of-the-art on OoDIS and generalizes to real unseen road obstacles, while retaining ID performance close to that of an ID-only detector across the three detector configurations we evaluate.
    \item We analyze what makes synthetic outliers effective for OOD object detection, and show 
    that performance is more sensitive to the quality of outlier annotations than to their amount and semantic content. 
\end{itemize}

\section{Related Work}
\label{sec:relatedwork}

\paragraph{\textbf{Outlier exposure and OOD detection.}}
Post-hoc scoring methods for closed-set classifiers~\cite{hendrycks2016baseline, 
liang2017enhancing, lee2018simple, sun2022out, ren2019likelihood, liu2020energy} 
rank the predictions a model already made. OE~\cite{hendrycks2018deep}   
has numerous extensions in classification~\cite{tao2023non, li2020background, zhang2023mixture}, 
~\cite{yu2024self, wang2023clipn, jiang2024negative, liu2024tag}. 
Among image-level OOD detection methods, Dream-OOD~\cite{du2023dream} and ~\citet{li2025synthesizing} learn a 
text-conditioned latent space and decode outliers from it as whole images. 
All of these decide whether an input image is anomalous, not where the anomaly is. 
Anomaly segmentation extends image-level 
OOD detection to a dense prediction task with 
established benchmarks, such as Fishyscapes~\cite{blum2019fishyscapes} and SegmentMeIfYouCan~\cite{chan2021segmentmeifyoucan}.
Early approaches scored pixels by uncertainty estimation or entropy maximization~\cite{oberdiek2020detection, chan2021entropy}. 
Recent methods~\cite{loiseau2024reliability, de2024placing} inpaint outliers with 
a diffusion model, UGainS~\cite{nekrasov2023ugains} obtains anomaly instances from pixel-level uncertainty, 
where supervision and evaluation remain at the pixel level, which does not transfer 
automatically to object detection. A dense predictor assigns a score to every pixel, 
so an anomaly always receives a prediction, whereas object detectors produce sparse predictions and may silently fail on 
unknown objects. 
\paragraph{\textbf{OOD object detection.}}
Detecting unknown objects at the instance level is shared with open-world 
object detection (OWOD)~\cite{joseph2021towards}, which incrementally 
learns unknowns it encounters~\cite{gupta2022ow, zohar2023prob}, whereas OOD object detection operates on a 
closed-set label space and detects objects that fall outside it as unknown. 
Most methods in this setting modify the detector with an auxiliary uncertainty 
branch~\cite{du2022vos}, a generalized objectness 
score~\cite{liang2023unknown}, a distance-based embedding head~\cite{du2022siren} or 
features from sensitivity-aware layers, trained on perturbed outliers~\cite{wilson2023safe}. 
These share a component that must be designed for OOD detection training 
and retained at inference time. 
VOS~\cite{du2022vos} samples virtual outliers from low-density regions of class-conditional Gaussians fitted to  
ID box features and passes them through the classification head, 
so they supervise how a detected box is classified, not whether it is proposed 
in the first place. 
SyncOOD~\cite{liu2024can} inpaints outliers and trains a binary classifier on box features of an off-the-shelf detector. 
In all of these, OOD detection applies to boxes the base detector already 
proposed, the OOD decision is made by an added scoring component and overlooked objects are not recovered.
Alternatively, zero-shot open-vocabulary object detectors (ZS-OVOD)~\cite{ilyas2024potential}  localize many street-scene obstacles~\cite{nekrasov2025oodis}, without OOD-specific training, 
however, their output is prompt-dependent and an unknown object is by definition 
not known ahead of time. 
Dream-Box~\cite{isaac2025dream} instead trains the region proposal network of Faster R-CNN on outliers, while its classification and box regression heads remain ID, and argues, that treating OOD objects as an additional class may not generalize.
That formulation is nevertheless established in image
classification~\cite{vernekar2019out} and semantic
segmentation~\cite{bevandic2022dense}. 
Unlike these methods, we treat outliers as detection targets that receive the 
same localization and classification supervision as any ID class, and find 
that the additional class transfers to real OOD objects.

\section{Method}
\begin{figure*}[t]
  \centering
   \includegraphics[width=\linewidth]{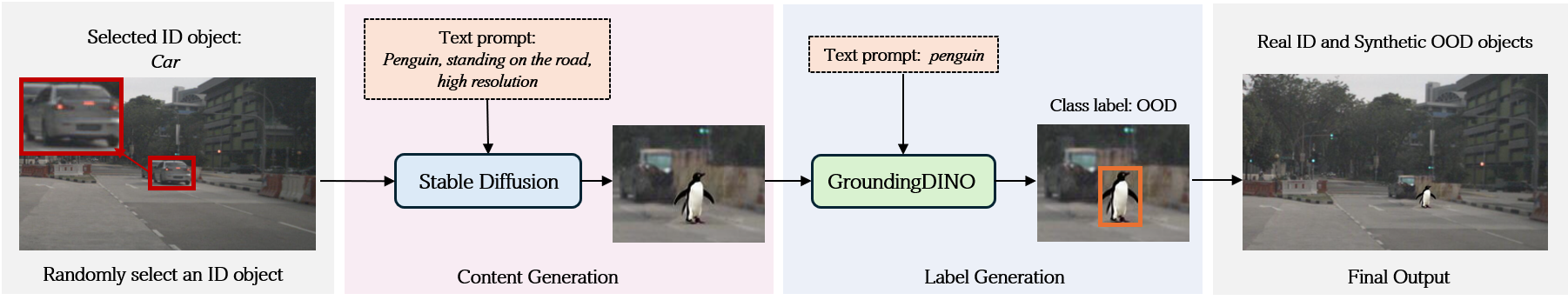}

   \caption{Overview of the synthetic outlier data generation process. The input image consists of ID objects, 
   \(X_\textit{ID}\), some of which are replaced with synthetic outliers using Stable Diffusion in the content generation step.
   The synthetic outliers are assigned to the OOD class, \(y_\textit{OOD}\), with a refined bounding box, \(b_\textit{OOD}\) using GDINO. 
   }
   \label{fig:main}
\end{figure*}

We propose SynOE-OD, that adapts an object detector to detect OOD and ID objects jointly. 
 We formulate the OOD object detection problem in \cref{prob_statement}. 
In \cref{outliers_finetuning}, \cref{assign_label} and \cref{fig:main} we describe 
(i) the synthetic outlier generation procedure based on Stable Diffusion for a given dataset  and
(ii) the annotations for synthetic outliers using GDINO. Lastly, \cref{sec:transfer} describes how the synthetic outliers are integrated into fine-tuning by transfer learning.

\subsection{Problem Statement}
\label{prob_statement}
For unified ID and OOD object detection, we consider a set of labeled ID 
objects \(X_{\textit{ID}}\) from a domain of interest, where each instance is a triplet \((x, b, y)\) of
an object \(x\), its bounding box \(b\) and a class label \(y\) from \(n\) 
known classes \(Y_{\text{ID}} = \{y_1, \ldots, y_n \}\). 
%\( (x,b,y) \in X_{\textit{ID}}\),  belongs to one of 
%the \( n \) known classes \( Y_{\text{ID}} = \{ y_1, \ldots, y_n \} \). 
The remaining objects %\( (x_\textit{OOD}, b_\textit{OOD}, y_\textit{OOD}) \in X_{\textit{OOD}} \), 
whose class is not contained in \(Y_{\textit{ID}}\) form the set \(X_{\textit{OOD}} \). %\(y_\textit{OOD}\). 
They are considered OOD and receive a single additional label \(y_{n+1}\), denoted by \(y_\textit{OOD}\). 
Given an object detector \(f_\theta\), the objective is to learn parameters \(\theta\) 
that localize the objects of \(X_\textit{ID}\) and classify them among the \(n\) known 
classes, and in the same scene localize the objects of \(X_\textit{OOD}\) 
and assign them to \(y_{n+1}\) rather than to the background.
The goal is to extend conventional object detection to the open-set 
setting.

\subsection{Synthetic outlier generation}
\label{outliers_finetuning}
\paragraph{\textbf{Outlier vocabulary.}} To obtain atypical object classes, we use ChatGPT \cite{chatgpt} 
to return a list of candidate objects that are unusual for a given dataset. 
Our application is autonomous driving perception, so we query ChatGPT to 
`\textit{give a list of objects that are unusual for a street scene}'. 
The returned list \(L_\textit{unusual} = \{l_{1}, l_{2}, \ldots, l_{k}\}\) 
contains \(k\) classes that fall into \textit{household items}, \textit{animals} and other \textit{miscellaneous} objects. 
The full list of outliers for street scenes is given in Appendix (\cref{full-voabulary}). 

\paragraph{\textbf{Scene editing.}}
\label{edit_scene}
We use Stable Diffusion \cite{runwayml2023inpainting} 
to inpaint unusual objects into street scenes, and modify two kinds of image regions. 
These regions are either locations of ID objects or free-space areas on the road. 
For the former, we randomly select an ID object \(x_\textit{ID}\) with 
bounding box \(b_\textit{ID}\) and class label \(y_\textit{ID}\), \((x_\textit{ID},b_\textit{ID}, y_\textit{ID}) \in X_\textit{ID}\). 
To keep the edit minimally invasive while integrating it smoothly into the scene, 
an enlarged image crop around the box \(b_\textit{ID}\) is passed to the model rather than the box alone, paired with a text prompt sampled 
from \(L_\textit{unusual}\) returned by ChatGPT. 
The input text-image pair thus consists of a random class \(l_\textit{unusual} \in L_\textit{unusual}\) 
and the cropped image region to generate the desired content. 
The generated content is then pasted back into the original image. 
The free-space inpainting follows the same procedure but differs in how the region is selected. 
The road regions are extracted from the ground truth masks, and multiple fixed-size boxes are placed
in random locations. The image content inside those boxes is then modified.

\subsection{\textbf{Annotation refinement}}
\label{assign_label}
\begin{wrapfigure}[11]{r}{0.5\linewidth}
  \centering
  \includegraphics[width=\linewidth]{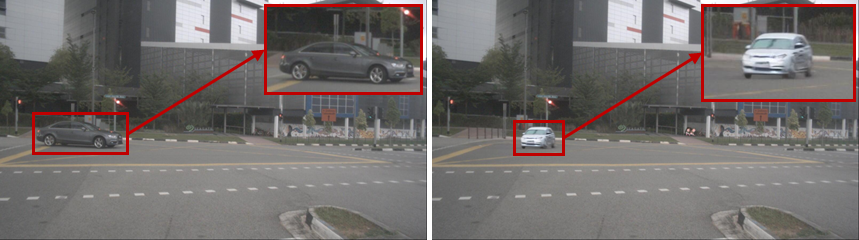}
  \caption{Left: original scene. Right: the same scene where the ID object
  \textit{car} was regenerated instead of an outlier.}
  \label{fig:ID_object}
\end{wrapfigure}
Text-guided inpainting does not always reliably generate the prompted content, 
a failure known as prompt neglect~\cite{manukyan2025hd}. 
What failures are possible depends on the region the inpainting mask covers. 
In free space, the prompted object may not be generated at all, leaving the region 
without an object. 
When an ID object is replaced, the model may instead regenerate a variant 
of that same ID object, 
and the mask region is filled with an edited ID object, see \cref{fig:ID_object}. 
In this case an object is generated and occupies the region, 
so neither a presence check nor an overlap test and similarity filter against \(b_\textit{ID}\), 
proposed in~\cite{liu2024can} catches it. 
Both cases if left uncorrected, introduce label noise, the first by annotating an empty region 
as OOD and the second by annotating an ID object as OOD.

We therefore employ GDINO for its exceptional zero-shot performance on semantically known classes, and distinguish three scenarios: 1) \textit{bounding box refinement} for correctly inpainted OOD objects,
2) \textit{label reassignment} when the inpainting reproduces the ID object it was meant to replace, 
and 3) \textit{bounding box removal} for empty inpaintings.
These three steps help us to obtain accurate bounding boxes and class labels, shown in \cref{fig:main}.
GDINO operates on image-text pairs \((B, l)\), where \(B\) is a crop around the original bounding box \(b_\textit{ID}\) 
of the replaced ID object and \(l\) is some text prompt. We limit the input image to a crop rather than
the full image, since on the full image the refinement is inconsistent and often misses the inpainted object.
To estimate the quality, we manually reviewed the updated annotations for a subset of the dataset.

\noindent
\textbf{Bounding box refinement} proceeds as follows: For each correctly inpainted OOD object the 
corresponding bounding box is refined by incorporating \(b_\textit{ID}\). 
We query GDINO with an image-text pair \((B_\textit{ID}, l_\textit{unusual})\), where \(B_\textit{ID}\) is the image region inside the original bounding box 
\(b_\textit{ID}\) and \(l_\textit{unusual} \in L_\textit{unusual}\) is 
the class sampled for inpainting.  
%\begin{equation}
%    \begin{aligned}
%        b_\textit{OOD} = \textit{GDINO}(B_\textit{ID}, l_\textit{unusual}).
%    \end{aligned}
%\end{equation}
The returned bounding box \(b_\textit{OOD}\) replaces the \(b_\textit{ID}\) 
and is assigned the class label \(y=y_{n+1}\).

\noindent
\textbf{Label reassignment} detects inpaintings that reproduced the original ID object and neglected the 
text prompt \(l_\textit{unusual} \in L_\textit{unusual}\), we query GDINO with the outlier category \(l_\textit{unusual}\) 
concatenated with the ID class label \(y_{\textit{ID}}\) of the replaced object, 
\begin{equation}
    l = l_\textit{unusual} \circ y_\textit{ID} \coloneqq l_\text{unusual}\, . \, y_\text{ID},
\end{equation}
e.g. ``penguin . car'', GDINO outputs the predicted class label \(y\): 
\begin{equation}
    y = \textit{GDINO}(b_\textit{ID}, l_\textit{unusual} \circ y_\textit{ID}).
\end{equation}
If the predicted class is \(y_\textit{ID}\), Stable Diffusion 
did not follow the prompt \(l_\textit{unusual}\) but regenerated a variant of the object it was
meant to replace. %See \cref{fig:ID_object} for an example. 
We revert these samples by restoring the original image content and retain the class label by setting \(y =y_\textit{ID}\). 
Reverted samples are not resampled, hence, the number of OOD instances is $11$\,\% lower than the inpainting attempts.

\noindent
\textbf{Bounding box removal} is performed in cases where GDINO returns no detections,
the inpainting carries no recoverable object. We treat such inpaintings as uninformative and discard the corresponding box \(b_\textit{ID}\), 
thereby excluding it from being assigned to any specific object class.

\subsection{\textbf{Transfer Learning for OOD detection}} \label{sec:transfer}
To optimize ID and OOD detection jointly, we start from object detectors pre-trained on large-scale datasets, e.g., Objects365 or medium-scale datasets 
such as COCO, and fine-tune them via transfer learning on our inpainted datasets that combine 
real and synthetic objects. The real objects preserve ID performance, whereas the 
synthetic outliers extend the detector to semantic concepts beyond the ID classes. 
Following \cref{prob_statement}, we add an \(n\)+\(1\)-st class bucket for OOD objects. 
Thereby, in our formulation the \(n\) ID classes are the set of object 
categories relevant to the original scenes, while objects outside this label space are recovered as OOD objects and are assigned to the \(n\)+\(1\)-st class.

\section{Experiments}

We describe the models, datasets and evaluation protocol in the experimental setup \ref{experimental_setup}, the quantitative 
results in \ref{results}, detailed ablations in \ref{detailed-experiments} and \ref{syn_effect} and implementation details in Appendix \ref{imp_details}. 

\subsection{Experimental setup}
\label{experimental_setup}
\paragraph{\textbf{Object Detection Models.}}
We apply SynOE-OD on both closed- and open-vocabulary object detectors: DINO-DETR (DD) \cite{zhang2022dino} which is vision-only 
and GDINO \cite{liu2024grounding}, which additionally conditions on text.
Both are fine-tuned on our inpainted datasets, which contain real ID objects alongside synthetic outliers.
For DINO-DETR (DD) we use two backbones, i.e., ResNet-50 \cite{he2016deep} and Swin-L \cite{liu2021swin}, 
each initialized from COCO-pretrained detector weights. 
To fine-tune GDINO we use a Swin-T \cite{liu2021swin} backbone initialized from the weights
pre-trained on O365, GoldG, GRIT, and V3Det.

\paragraph{\textbf{Datasets and Evaluation.}}
\label{datasets}
We use three street scene datasets: NuImages~\cite{Caesar2020nuscenes}, Cityscapes~\cite{Cordts2016Cityscapes} and BDD100K~\cite{Yu2020BBD100k}.
NuImages (NU) is our primary training set and supplies the ID classes. 
Cityscapes (CS) and BDD100K (BDD) are added to the training data to test whether the OOD results are particularly influenced by domain gaps between OOD and ID data. 
We evaluate OOD object detection on RoadObstacle (RO), 
RoadAnomaly (RA), Fishyscapes (FS), included in OoDIS, and on LostAndFound (LF) \cite{pinggeralost}. 
For evaluating ID performance, we use the held out validation set from NuImages. 
Since we are specifically interested in the detection capability of object detectors, we align the evaluation with standard object detection practice \cite{padilla2020survey}, 
and adopt the evaluation protocol of OoDIS. Thus, we report COCO-style 
Average Precision (AP\textsubscript{50..95}, AP\textsubscript{50}, AP\textsubscript{75}, AP\textsubscript{S,M,L}). 
FPR95 and AUROC are standard image-level OOD detection metrics~\cite{hendrycks2016baseline, liang2017enhancing, yang2024generalized} that require separate ID and
OOD sets and are not applicable in our case.

\paragraph{\textbf{Augmented Datasets for Fine-tuning.}}
Following the generation procedure described in \cref{outliers_finetuning}, we augment the base datasets by inserting synthetic outliers, while retaining the original ID annotations,
illustrated in the Appendix \cref{fig:datasets}. 
For NuImages this yields 10{,}702 images
carrying 82{,}197 ID and 15{,}604 OOD annotations, so outliers account for
\(16.0\,\%\) of the labelled objects. For Cityscapes, 2{,}975 images with
44{,}621 ID and 5{,}568 OOD annotations, whereas for BDD100k, 2{,}000 images with 22{,}202 ID
annotations and 3{,}109 OOD annotations. These augmented datasets are what the object detectors are fine-tuned on.

\paragraph{\textbf{Baselines.}}
We compare against three kinds of baseline, summarized in the Appendix \cref{tab:baselines}.
The first are the published state-of-the-art OoDIS methods, ZS-OVOD~\cite{ilyas2024potential} and UGainS~\cite{nekrasov2023ugains}.
The second is a training-based OOD method, VOS~\cite{du2022vos}, which we train
on NuImages and evaluate through the same OoDIS protocol. 
The third is the models that differ from SynOE-OD in not having seen synthetic
outliers.

\begin{table}[t]
\centering
\small
\caption{Results on the OoDIS benchmark. Size specific AP is mentioned in the Appendix \cref{size-ap}.}
\label{tab:benchmark}
\setlength{\tabcolsep}{4pt}
\begin{tabular}{l ccc ccc ccc}
\toprule
& \multicolumn{3}{c}{\textbf{RoadObstacle}}
& \multicolumn{3}{c}{\textbf{RoadAnomaly}}
& \multicolumn{3}{c}{\textbf{Fishyscapes}} \\
\cmidrule(lr){2-4}\cmidrule(lr){5-7}\cmidrule(lr){8-10}
\textbf{Method}
& AP\textsubscript{50:95} & AP\textsubscript{50} & AP\textsubscript{75}
& AP\textsubscript{50:95} & AP\textsubscript{50} & AP\textsubscript{75}
& AP\textsubscript{50:95} & AP\textsubscript{50} & AP\textsubscript{75} \\
\midrule
ZS-OVOD & 48.5 & 77.6 & 52.2 & 22.7 & 33.3 & 23.3 & 22.7 & 33.4 & 24.5 \\
UGainS   &  8.1 & 11.6 &  8.9 & 11.7 & 17.5 & 12.1 & 15.6 & 24.3 & 18.3 \\
\midrule
\rowcolor{gray!20} SynOE-OD (DD:ResNet50)    & 48.8 & 79.4 & 49.4 & 26.9 & 35.8 & 29.0 & 35.3 & 55.3 & 37.9 \\
\rowcolor{gray!20} SynOE-OD (DD:Swin-L) & \textbf{57.6} & \textbf{87.5} & \textbf{61.7} & 25.7 & 32.1 & 27.2 & 35.9 & 49.5 & 40.7 \\
\rowcolor{gray!20} SynOE-OD (GDINO:Swin-T) & 55.6 & 85.1 & 60.2 & \textbf{29.4} & \textbf{37.3} & \textbf{31.5} & \textbf{39.2} & \textbf{55.8} & \textbf{42.2} \\
\bottomrule
\end{tabular}
\end{table}

\subsection{Main Results}
\label{results}

\paragraph{Comparison on the OoDIS benchmark.}
OoDIS scores anomalies at the instance level, which currently makes it the only publicly-available benchmark where localizing OOD objects is measured. We submitted all three detectors fine-tuned on inpainted NuImages and Cityscapes (\cref{tab:benchmark}).
SynOE-OD establishes new state-of-the-art performance on all three test sets, by over $16$ percentage points (pp) on Fishyscapes, $9$\,pp  on RoadObstacle and $7$\,pp on RoadAnomaly.  
The absolute performance across the three benchmarks is not uniform, which is expected as the datasets 
differ in domain and in the type of OOD objects they contain. 
RoadAnomaly images are collected from the web rather than recorded by a camera mounted on a vehicle, so the viewpoint and object scale are very different, which makes it furthest from our training domain. RoadObstacle and 
Fishyscapes have a forward-looking viewpoint consistent with NuImages and contain small OOD objects scattered across the 
street, the setting our free-space inpainting reproduces directly, 
and the numbers reflect this. On RoadObstacle, AP\textsubscript{50} reaches $87.5$\%, so obstacles far away and in isolation are reliably detected. RoadAnomaly instead contains dense groups of adjacent anomalies, e.g., images with a flock of sheep, whereas our training images carry at most three outliers.

\paragraph{Comparison to trained OOD baseline.}
For VOS, no official results have been reported on OoDIS.
% has no entry on OoDIS
For a fair comparison we retrain it on original NuImages and evaluate it under the OoDIS protocol.
 We also include an ID-only baseline trained on NuImages, same as VOS, whose predictions are remapped to the OOD class and evaluated class-agnostically (\cref{tab:vos}).
 SynOE-OD (DD:ResNet50) is fine-tuned on inpainted NuImages and compared against these two baselines.
SynOE-OD outperforms VOS on all three benchmarks. %by $46.2$, $23.0$ and $27.7$\,pp on RoadObstacle, RoadAnomaly and LostAndFound.
 VOS detects very few OOD objects but has a competitive ID mAP, and so does the ID-only baseline.
% The reasons for its high ID score are that its object detection head does 
% not carry an additional class, and that it is trained roughly $4\times$ longer than SynOE-OD, \SA{which gives it more ID supervision compared to SynOE-OD and the ID-only baseline.} %following their published schedule. 
 The ID-only model outperforms VOS on every benchmark and recovers far more objects in RoadAnomaly with five times as many false positives. Thus, one of these barely proposes boxes and the other proposes far too many. Supervising outliers as detection targets, as in our method, is what balances precision and recall at once.
%\SA{added ID only now, write text later, synoe is epoch 3 here change to epoch 2}

\begin{table}[h]
  \caption{Comparison to feature-space outlier synthesis and ID-only baseline. All three methods see the same ID data during training. 
  Recall is computed over all ground-truth OOD instances %($557$ / $731$ / $1928$).
  }
  \label{tab:vos}
  \centering
  \footnotesize
  \setlength{\tabcolsep}{4pt}
  \begin{tabular}{llcccccc}
    \toprule
    \textbf{Method} & \textbf{Benchmark} & \textbf{AP\textsubscript{50:95}}
      & \textbf{TP} & \textbf{FP} & \textbf{FN} & \textbf{Recall}
      & \textbf{ID} \\
    \midrule
    \multirow{3}{*}{ID-only}
  & RoadObstacle  & 11.3 & 142 & 144  & 415  & 25.5
    & \multirow{3}{*}{\textbf{42.2}} \\
  & RoadAnomaly   &  2.4 & 445 & 6772 & 286  & \textbf{60.9} & \\
  & LostAndFound  &  5.4 & 553 & 3932 & 1375 & 28.7 & \\
\midrule
    \multirow{3}{*}{VOS}
      & RoadObstacle  & 1.4 & 16   & 17   & 541  & 2.9
        & \multirow{3}{*}{42.0} \\
      & RoadAnomaly   & 2.2 & 291  & 2359 & 440  & 39.8 & \\
      & LostAndFound  & 2.1 & 334  & 921  & 1594 & 17.3 & \\
    \midrule
    \multirow{3}{*}{SynOE-OD (ours)}
      & RoadObstacle  & \textbf{47.6} & 474  & 471  & 83  & \textbf{85.2}
        & \multirow{3}{*}{40.6} \\
      & RoadAnomaly   & \textbf{25.2} & 419  & 1222 & 312 & 57.3 & \\
      & LostAndFound  & \textbf{29.8} & 1245 & 2565 & 683 & \textbf{64.6} & \\
    \bottomrule
  \end{tabular}
\end{table}

\subsection{Results across object detectors and datasets}
\label{detailed-experiments}
\paragraph{Generality across detectors and backbones.}
%OOD object detection should not be tied to a single detector,
 Closed-set detectors are trained only on ID classes leaving OOD objects without an explicit detection target, causing them to be treated as background. %assign OOD objects to the background, as their training provides no alternative outcome. 
We therefore test whether SynOE-OD supports OOD detection in closed-set detectors and also benefits OVODs. SynOE-OD (DINO-DETR), fine-tuned on inpainted NuImages with both ResNet-50 and Swin-L backbones, generalizes %from synthetic outliers 
to real OOD objects in the test benchmarks (\cref{tab:models}), although the model never encounters real OOD objects during training. Swin-L performs  best when the test domain closely matches the training domain (RoadObstacle) and worst were it does not (RoadAnomaly), which is consistent with a larger backbone fitting the training distribution more.
The comparison for GDINO is particularly important, since its zero-shot formulation already detects unseen objects without any task-specific training, and therefore together with its prompt ensemble serves as a strong baseline. 
SynOE-OD (GDINO) improves the performance by $6$, $7$ and $3$\,pp on RoadObstacle, RoadAnomaly and LostAndFound.
%and sees only synthetically generated outliers. 
Synthetic outliers supply a training target for OOD objects that prompting alone does not provide.
  %spans at different levels, while the reduction for all three is at most $\sim$2\,pp. 
Our method creates OOD detection capability in a closed-set detector and improves it in OVODs, with a small reduction of at most $\sim$2\,pp in ID performance.
%The reduction in ID performance across all three detectors is at most $\sim$2\,pp.

\begin{table}[tb]
\caption{OOD and ID object detection results with three detectors compared against GDINO zero-shot and prompt ensemble~\cite{ilyas2024potential}. ID$_\text{base}$ is the same detector without the OOD class and trained on original NuImages.}
\label{tab:models}
\centering
\small
\setlength{\tabcolsep}{8pt}
\begin{tabular}{lccc|cc}
\toprule
\textbf{Method} & RO & RA & LF & ID & ID$_\text{base}$ \\
\midrule
GD zero-shot       & 48.3 & 19.5 & 26.4 & --- & --- \\
GD prompt ensemble  & 48.5 & 22.7 & ---  & --- & --- \\
\midrule
SynOE-OD (DD:R50)    & 48.7 & 26.0 & \textbf{32.0} & 38.8 & 40.4 \\
SynOE-OD (DD:Swin-L) & \textbf{56.5} & 24.5 & 27.0 & 49.1 & 50.8 \\
SynOE-OD (GDINO:Swin-T) & 54.3 & \textbf{29.3} & 29.2 & 45.8 & 47.2 \\
\bottomrule
\end{tabular}
\end{table}

\paragraph{Effect of the text prompt.} 
OVODs such as GDINO are conditioned on text; consequently, the name assigned to the OOD class can have a significant effect on its OOD detection performance.
%so the name given to the OOD class is rather a design choice. 
%We first use the same class name as for the closed-set setting. 
OOD objects cannot be enumerated beforehand, hence the class name has to be generic, and unlike prompt-based methods, we keep it fixed at inference time.
GDINO encodes class names with a BERT text encoder, and we compare how it represents different class names (\cref{tab:bert}). 
Specifically, we measure their cosine similarity to the embedding of \textit{obstacle},
%tokenizes different class names and how close their embeddings lie to . 
%We compare its cosine similarity to the class name \textit{obstacle}, 
due to its conceptual relevance to out-of-distribution objects (\cref{tab:cs}); the other candidates are listed for context.
The class name \textit{OOD} carries no meaning: BERT splits it into two subword tokens and its embedding lies far from that of \textit{obstacle}. Therefore, the model has to learn a new concept from scratch. 
As a result, we use \textit{anomaly}, which improves RoadAnomaly by $5.5$\,pp (\cref{tab:anomaly}) while LostAndFound loses $1.7$\,pp, with ID performance almost unaffected. 
Training with \textit{'anomaly'} also starts at a
lower loss, $10.06$ against $10.36$, consistent with a better-initialised 
text embedding. 
Hence, a semantically meaningful class name provides a stronger initialization and improves OOD object detection.

\begin{table}[h]
\caption{Effect of the OOD class name on SynOE-OD (GDINO).}
\label{tab:bert}
\centering\footnotesize
\begin{subtable}[t]{0.54\linewidth}
\caption{AP\textsubscript{50..95} score for GDINO fine-tuned with two different prompts with a learning rate of \num{2e-4}. The ID mAP is unaffected ($47.3$ against $47.4$).}
\label{tab:anomaly}
\centering
\setlength{\tabcolsep}{2pt}
\begin{tabular}{lccc}
\toprule
\textbf{Class name} & \textbf{RO} & \textbf{RA} & \textbf{LF} \\
\midrule
\texttt{OOD}     & 53.0 & 21.4 & 27.4 \\
\texttt{anomaly} & 53.2\dlt{+0.2} & 26.9\dlt{+5.5} & 25.7\dlt{-1.7} \\
\bottomrule
\end{tabular}
\end{subtable}\hfill
\begin{subtable}[t]{0.42\linewidth}
\caption{BERT tokens and cosine similarity to the \textit{obstacle} class name.}
\label{tab:cs}
\centering
\setlength{\tabcolsep}{2.5pt}
\begin{tabular}{lcc}
\toprule
\textbf{Class prompt} & \textbf{Tokens} & \textbf{cos} \\
\midrule
\texttt{OOD}     & \texttt{['o','\#\#od']} & 0.21 \\
\texttt{unknown} & \texttt{['unknown']} & 0.57 \\
%\texttt{debris}  & 1 & 0.66 \\
\texttt{object}  & \texttt{['object']} & 0.71 \\
\texttt{anomaly} & \texttt{['anomaly']} & 0.74 \\
\bottomrule
\end{tabular}
\end{subtable}
\end{table}

\paragraph{Mitigating the domain gap with ID data variety.}
There is a significant domain gap between the NuImages training set and the evaluation benchmarks. We ask whether more variety in the ID training data can mitigate this gap.
Cityscapes is particularly relevant to LostAndFound, as both datasets were recorded in the same country and with the same camera setup. BDD100K in contrast shares its acquisition domain with none of the benchmarks, adding it therefore tests whether generic appearance diversity is beneficial.
Following the same inpainting procedure, we incorporate Cityscapes and 
BDD100K by only replacing ID objects, and add them both in the training set (\cref{tab:scale-ID}). 
On LostAndFound, adding Cityscapes narrows the domain gap and increases the performance for SynOE-OD  by $3.3$\,pp (DD:Swin-L) and $4.3$\,pp (GDINO), and adding BDD100k adds another point for both detectors, whereas ResNet50 does not benefit.
% yields a further point more for both detectors.
The performance on RoadObstacle and RoadAnomaly essentially stays constant.
ID performance decreases by a point for each detector, which we attribute to evaluating on a validation set exclusively drawn from NuImages.
Hence, SynOE-OD mitigates the domain gap mainly when the ID training data matches the target domain, while generic diversity contributes marginally.

\begin{table}[h]
\centering
\small
\caption{Effect of domain diversity in the training set. The parentheses show the difference to when trained on NuImages only.
}
\label{tab:scale-ID}
\setlength{\tabcolsep}{3.5pt}
\begin{tabular}{llcccc}
\toprule
\textbf{Method} & \textbf{Training data} & RO & RA & LF & NU \\
\midrule
\multirow{2}{*}{\makecell{SynOE-OD\\(DD:R50)}}
  & + CS       & 48.8\dlt{+0.0} & 26.1\dlt{+0.1} & 31.2\dlt{-1.3} & 37.4\dlt{-1.4} \\
  & + CS + BDD & 49.1\dlt{+0.3} & 26.0\dlt{+0.0} & 31.9\dlt{-0.6} & 37.7\dlt{-1.1} \\
\midrule
\multirow{2}{*}{\makecell{SynOE-OD\\(GD:Swin-T)}}
  & + CS       & 55.6\dlt{+1.3} & 28.2\dlt{-1.1} & 33.5\dlt{+4.3} & 45.0\dlt{-0.8} \\
  & + CS + BDD & 55.0\dlt{+0.7} & 27.5\dlt{-1.8} & 34.7\dlt{+5.5} & 44.8\dlt{-1.0} \\
\midrule
\multirow{2}{*}{\makecell{SynOE-OD\\(DD:Swin-L)}}
  & + CS       & 57.6\dlt{+1.1} & 25.2\dlt{+0.7} & 30.3\dlt{+3.3} & 48.5\dlt{-0.6} \\
  & + CS + BDD & 56.4\dlt{-0.1} & 24.4\dlt{-0.1} & 31.3\dlt{+4.3} & 48.3\dlt{-0.8} \\
\bottomrule
\end{tabular}
\end{table}

\begin{wraptable}[15]{r}{0.4\linewidth}
\caption{Per class ID performance in AP\textsubscript{50..95}. The annotation count is in the Appendix \cref{tab:ID-perclass-ann}.}
\label{tab:ID-perclass-small}
\vspace{-\intextsep}
\centering\small
\setlength{\tabcolsep}{2pt}
\begin{tabular*}{\linewidth}{@{\extracolsep{\fill}}lcc@{}}
\toprule
Class & ID baseline & SynOE-OD \\
\midrule
car        & 58.3 & 57.1\dlt{-1.2} \\
pedestrian & 42.3 & 41.4\dlt{-0.9} \\
truck      & 42.8 & 40.6\dlt{-2.2} \\
motorcycle & 47.3 & 44.9\dlt{-2.4} \\
bus        & 42.4 & 41.7\dlt{-0.7} \\
bicycle    & 35.8 & 32.3\dlt{-3.5} \\
trailer    & 16.6 & 13.4\dlt{-3.2} \\
\midrule
mAP        & 40.8 & 38.8\dlt{-2.0} \\
\bottomrule
\end{tabular*}
\end{wraptable}
\paragraph{ID/OOD trade-off.}
Joint ID and OOD detection can affect ID performance in two ways: the additional OOD class competes with the ID classes for predictions, and replacing ID objects with outliers removes training annotations for the affected classes. 
To quantify the combined impact of these effects, we compare class-wise ID performance against a baseline without an OOD class and with all of its original ID annotations retained. As shown in \cref{tab:ID-perclass-small}, SynOE-OD retains its ID mAP within $2.0$\,pp of the baseline. The performance reduction across classes is not uniform and is dependent on how frequently each class appears in the dataset. The underrepresented classes \textit{motorcycle}, \textit{bicycle} and \textit{trailer} exhibit the largest decreases, whereas the abundant classes \textit{car} and \textit{pedestrian} decline only by about a point. However, \textit{bus} is minimally affected, despite its rarity, as none of its objects are replaced during outlier generation. Nearly all of the reduction is concentrated in classes that are both rare and replaced. 
SynOE-OD enables joint ID and OOD object detection while preserving about $95$\,\% of the baseline's ID performance.

\subsection{What makes synthetic outliers effective}
\label{syn_effect}

\paragraph{Synthetic outlier vocabulary.}
Here we measure how far SynOE-OD generalizes to semantic concepts that are not part of its outlier vocabulary.
The outlier vocabulary is proposed by an LLM (ChatGPT) without any reference to the 
test sets, thus, some of its classes inevitably appear in them. 
Out of the 90 classes in the outlier vocabulary, 20 occur in the test sets. 
We regenerate the training set by removing those 20 classes, so their inpaintings are undone. 
As a counterpart we remove 20 non-overlapping classes, which are contained in 
none of the test sets, i.e., vacuum, chair, lamp, printer etc.\ \cref{tab:oe-voc}.  
For SynOE-OD (DINO-DETR), removing the 20 overlapping classes decreases performance at most by $1.5$\,pp, 
whereas removing 20 non-overlapping classes decreases %$0.4$, $0.5$ and $0.1$\,pp
less than $1$\,pp on the same benchmarks. 
This magnitude is, however, model dependent, GDINO drops by $4.1$\,pp on LostAndFound. 
This overall decrease in performance reflects the reduced diversity of outliers in the training set rather than the absence of overlapping concepts. SynOE-OD therefore generalizes to unseen concepts and benefits from diversity in the outlier vocabulary.

\begin{table}[tb]
\centering
\caption{Full vocabulary is the standard dataset with 15{,}604 outlier annotations.
The 20 overlapping classes removal reduces the annotations to 12{,}863 and the non-overlapping 
ones to 12{,}832.}
\label{tab:oe-voc}
\small
\setlength{\tabcolsep}{6pt}
\begin{tabular}{llccc}
\toprule
\textbf{Model} & \textbf{Annotation}  & RO & RA & LF \\
\midrule
\multirow{3}{*}{SynOE-OD (DD:R50)}
 & baseline --- full vocabulary             & 48.7 & 26.0 & 32.0 \\
 & 20 overlapping classes removed & 47.2 & 25.8 & 30.5 \\
 & 20 non-overlapping classes removed & 48.3 & 25.9 & 31.5 \\
\midrule
\multirow{3}{*}{SynOE-OD (GD:Swin-T)}
  & baseline --- full vocabulary             & 54.3 & 29.3 & 29.2 \\
  & 20 overlapping classes removed  & 53.8 & 28.4 & 25.6 \\
  & 20 non-overlapping classes removed  & 53.8 & 27.7 & 25.1 \\
\bottomrule
\end{tabular}
\end{table}

\paragraph{Proportion of synthetic outliers.}
The proportion of synthetic outliers determines the balance between ID and OOD performance, which are optimal at different levels. 
How many outliers are needed is a key parameter of SynOE-OD, and we measure how much ID performance is exchanged for them.
We vary the fraction of the training data that carries synthetic outliers while keeping the image count fixed at $10{,}702$ (\cref{tab:proportion}).
A quarter of the outliers already delivers most of the OOD performance, with ID mAP essentially unchanged. 
\begin{wraptable}[13]{r}{0.45\linewidth}
\caption{Effect of synthetic outliers proportion. 
The outlier percentage shows the inpainted images retained. This corresponds to $3.3$\%, $6.5$\%, $9.2$\% and $13.2$\% 
of the annotations being OOD and every other image is
the original scene with its ID annotations.}
\label{tab:proportion}
\vspace{-\intextsep}
\centering\footnotesize
\setlength{\tabcolsep}{4pt}
\begin{tabular*}{\linewidth}{@{\extracolsep{\fill}}lcccc@{}}
\toprule
Outliers & RO & RA & LF & NU \\
\midrule
$0$\,\% (ID-only)   & 12.0 &  2.4 &  5.2 & 40.4 \\
$25$\,\%  & 46.4 & 26.1 & 29.9 & 40.0 \\
$50$\,\%  & 48.2 & 26.2 & 30.4 & 39.9 \\
$70$\,\%  & 48.1 & 25.5 & 31.0 & 39.2 \\
$100$\,\% & 48.7 & 26.0 & 32.2 & 38.7 \\
\bottomrule
\end{tabular*}
\end{wraptable}
RoadAnomaly does not improve beyond this point, 
whereas RoadObstacle reaches its plateau at half the outliers,   
and LostAndFound continues to improve up to the full set.
ID performance decreases monotonically by up to $2$\,pp.
Most of the OOD capability is achieved by using only a quarter of the outliers at no measurable ID cost. 
Preserving ID performance to maximum favors a smaller proportion of outliers, whereas maximizing OOD performance requires a larger proportion of outliers.

\paragraph{Annotation refinement.}
The annotation refinement step in our method is crucial and affects both ID and the OOD performance of SynOE-OD, see appendix \cref{tab:unrefined}. 
Without bounding box refinement, OOD AP drops significantly 
while ID performance stays almost unchanged $33.7$\,\% against $34.5$\,\%. %\cref{tab:unrefined}. 
Label reassignment identifies 1{,}307 samples for 
\textit{car}, that are regenerated as a \textit{car}, and restoring 
them increases \textit{car} AP from $51.0$\,\% to $54.7$\,\%. 
Without restoring these regions, the generated outliers contain modified ID objects and OOD class is contaminated with ID objects. Therefore, annotation refinement is essential to maintain a clear separation between ID and OOD supervision and to prevent training ID objects as OOD.

\section{Conclusion}
We address out-of-distribution object detection in street scenes and present SynOE-OD, which jointly detects ID and
OOD objects with a single detector and requires no additional auxiliary branch at inference. Synthetic outliers are
inpainted into street scenes with Stable Diffusion, their annotations are refined with an open-vocabulary detector, and
the detector is fine-tuned with one additional class. Evaluations on the OoDIS benchmark show that our approach
substantially surpasses existing zero-shot OVOD baselines, confirming that zero-shot detection alone is insufficient
for OOD object detection, and that scoring based methods alone cannot recover missed OOD objects. These improved results hold consistently across closed-set and open-vocabulary detectors and training
datasets, while the cost in ID performance remains small. Synthetic outliers prove effective mainly through accurate
annotations and semantic diversity, and the resulting detection capability generalizes to real, unseen road
obstacles.

\newpage

\bibliography{main}
\bibliographystyle{unsrtnat}

\clearpage
\appendix
%\section{Appendix}
\section{Implementation details}
\label{imp_details}

\begin{figure}[h]
  \centering
  \begin{subfigure}[b]{\textwidth}
  \centering
   \includegraphics[width=0.9\linewidth]{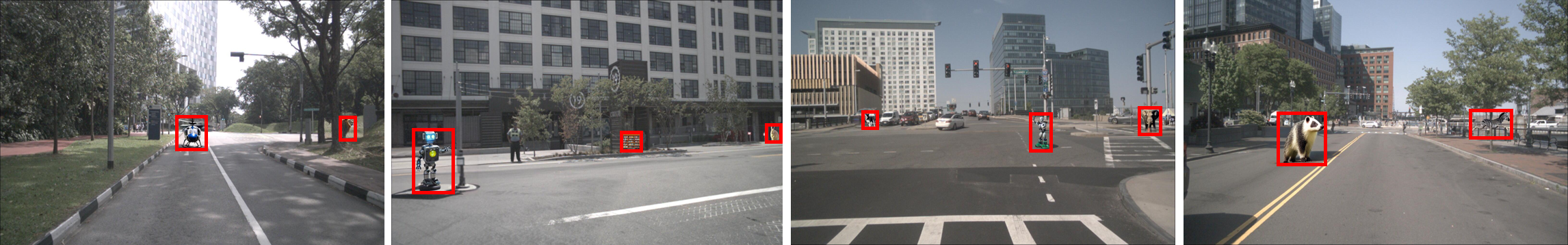}
    \end{subfigure}
    \begin{subfigure}[b]{\textwidth}
    \centering
   \includegraphics[width=0.9\linewidth]{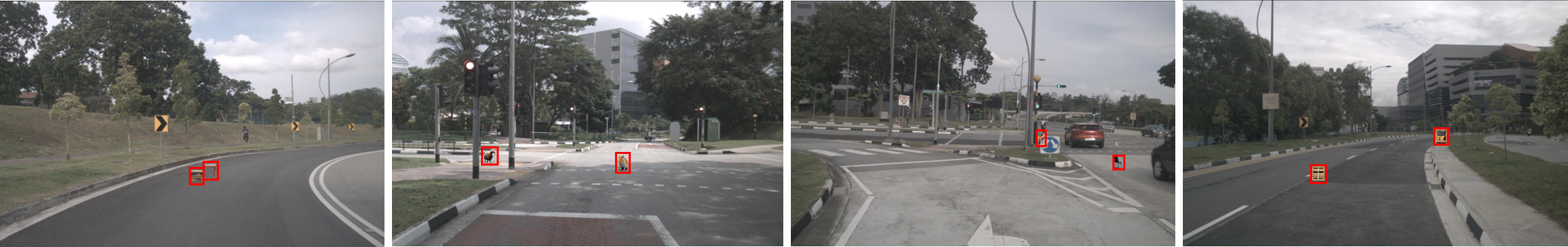}
    \end{subfigure}
   \caption{Illustration of the inpainted synthetic outliers on NuImages. 
   The first row shows the ID objects being replaced with OOD objects of different sizes while 
   maintaining a minimum distance. The second row shows objects inpainted directly 
   into the free-space i.e., the road region. 
   }
   \label{fig:datasets}
\end{figure}

\paragraph{Stable Diffusion.}
The inpainting quality is greatly affected by the size of the crop region. We use three different sizes, i.e., 
for small objects $128\times128$ pixels, while for medium sized $256\times256$ pixels and for large objects $512\times512$ pixels. However, the Stable Diffusion model has a base resolution of $512\times512$ pixels; thus, the input to the model is always resized to $512\times512$ beforehand.
The number of inpainted objects for each image is selected randomly and often times there is more than one inpainted OOD object.
Before inpainting multiple objects in a single image, we make sure that there is a minimum distance between the cropped regions, such that the inpaintings are spatially separated. The minimum distance is based on the largest crop size, i.e.\,, $512\times512$ pixels, that enforces a minimum center distance of 512 pixels. The resulting inpaintings are shown in  \cref{fig:datasets}.

\paragraph{Object detectors.} 
All detectors are fine-tuned on 3 GPUs with AdamW, a cosine-anealing learning 
schedule and an exponential moving average of the weights.
For DINO-DETR with a ResNet-50 backbone, we kept the backbone stages 
frozen and the base learning-rate is set to $2\times10^{-4}$ with a linear 
warm-up for the initial 250 iterations with a starting factor $0.001$, the batch 
size is 6 per GPU, the EMA momentum is $5\times10^{-3}$.
DINO-DETR with Swin-L and GDINO with Swin-T are fine-tuned with backbone stages 
unfrozen, a base learning rate of $1e-4$ and no warm-up. Swin-L uses a batch size 
of 2 per GPU, an EMA momentum $2\times10^{-3}$ and GDINO uses a batch size of 6 
per GPU with $5\times10^{-3}$ EMA momentum. 
The unfrozen backbones and as well as the language model of GDINO are trained at 
$0.1\times$ the base learning rate. For DINO-DETR with ResNet-50, the sampling 
offsets of the deformable attention are also trained $0.1\times$ the base 
learning rate.

\paragraph{Datasets.}
The three ID datasets differ in acquisition: NuImages is derived from NuScenes 
with 2D labels, Cityscapes was captured in German cities in daylight and dry 
weather, and BDD100K was recorded in the USA across varied weather and lighting conditions.    
For NuImages we use the front camera images, which are consistent in camera view with the other two datasets. 
From BDD100K, we randomly selected 2K images, to prevent it from dominating the combined training set, 
and from Cityscapes 2{,}975 finely annotated training images. 
In analogy to the NuScenes benchmark, we use its six vehicle classes and the pedestrian class, 
which are \textit{bicycle}, \textit{bus}, \textit{car}, \textit{motorcycle}, 
\textit{trailer}, \textit{truck} and \textit{pedestrian}. 
We exclude \textit{cones}, \textit{barriers} and \textit{construction}, those are not typical ID street scene objects and are treated as OOD in RoadAnomaly.

\paragraph{Baselines.}
\small
\Cref{tab:baselines} groups the baselines by what each one controls for. The
ID-only model has no OOD class, so every prediction is remapped to that category
before.
\begin{table}[h]
\caption{Baselines and what each accounts for.}
\label{tab:baselines}
\vspace{-\baselineskip}
\centering\footnotesize
\setlength{\tabcolsep}{3pt}
\begin{tabular}{ll}
\toprule
\textbf{Baseline} & \textbf{Comparison} \\
\midrule
ZS-OVOD & OoDIS  \\
UGainS  & OoDIS  \\
VOS & training-based OOD \\
ID-only (ours) & effect of inpainted data \\
\bottomrule
\end{tabular}
\end{table}

\paragraph{Fine-tuning Configuration.}
We vary the number of frozen backbone stages, the learning-rate schedule and the
transformer dropout to find the setup whose OOD performance is most stable across epochs (\cref{tab:training-parameters}, \cref{tab:swinl-training}).
The fully frozen ResNet50 backbone with cosine annealing is more stable across epochs, whereas for Swin-T and Swin-L unfrozen is better.

\begin{table}[h]
\caption{Backbone freezing and learning-rate warm-up for DINO-DETR with Swin-L on inpainted NuImages. The standard
configuration is the mean over three runs; each variant changes one setting.}
\label{tab:swinl-training}
\centering
\small
\begin{tabular}{lccc}
\toprule
\textbf{Configuration} & \textbf{RO} & \textbf{RA} & \textbf{LF} \\
\midrule
Standard (no warmup, unfrozen) & {0.578} & {0.264} & {0.306} \\
\midrule
+ LinearLR warmup         & 0.563 & 0.260 & 0.266 \\
\texttt{frozen\_stages=2}   & 0.575 & 0.241 & 0.275 \\
\texttt{frozen\_stages=1}   & 0.574 & 0.232 & 0.274 \\
\bottomrule
\end{tabular}
\end{table}

\begin{table}[h]
\centering
\small
\caption{Ablation grid - ID and OOD metrics averaged over seeds (24, 42). `Frozen' refers to the backbone stages. The dropout is added to the transformer encoder, decoder. The model trained is DINO-DETR with ResNet50 on NuImages.}
\setlength{\tabcolsep}{6pt}
\begin{tabular}{llcccc}
\toprule
\textbf{Config} & \textbf{Epoch} & \textbf{ID mAP} & \textbf{RO} & \textbf{RA} & \textbf{LF} \\
\midrule
\multirow{2}{*}{A (frozen=1, flat LR)}
 & 1 & 32 & 47 & 26 & 30 \\
 & 3 & 38 & 43 & 22 & 21 \\
\midrule
\multirow{2}{*}{B (frozen=4, flat LR)}
 & 1 & 31 & 47 & 26 & 32 \\
 & 3 & 36 & 44 & 23 & 23 \\
\midrule
\multirow{2}{*}{C (frozen=4, cosine+warmup)}
 & 1 & 30 & 47 & 26 & 32 \\
 & 3 & 36 & 46 & 24 & 25 \\
\midrule
\multirow{2}{*}{D (frozen=4, cosine-only)}
 & 1 & 31 & 46 & 26 & 30 \\
 & 3 & 37 & 46 & 23 & 23 \\
\midrule
\multirow{2}{*}{\shortstack[l]{E (frozen=4, dropout=0.1,\\ warmup+cosine)}}
 & 1 & 29 & 47 & 25 & 35 \\
 & 3 & 35 & 47 & 24 & 27 \\
\midrule
\multirow{2}{*}{\shortstack[l]{F (frozen=4, cosine-only,\\ drop=0.1)}}
 & 1 & 31 & 47 & 25 & 33 \\
 & 3 & 36 & 46 & 23 & 23 \\

\bottomrule
\end{tabular}
\label{tab:training-parameters}
\end{table}

\paragraph{GDINO learning rate.}
The learning rate for GDINO trades OOD performance against ID perfromance. 
We show that a higher learning rate is better for ID performance while worse for OOD detection. See (\cref{tab:lr}).

\begin{table}[H]
\caption{Doubling the learning rate raises ID mAP and lowers OOD AP.}
\label{tab:lr}
\centering
\small
\begin{tabular}{llcccc}
\toprule
\scriptsize
\textbf{lr} & \textbf{Test/Val set} & ep1 & ep2 & ep3 & ep4 \\
\midrule
1e-4 & RO     & {0.560} & 0.538 & 0.532 & 0.537 \\
1e-4 & RA     & {0.298} & 0.284 & 0.271 & 0.258 \\
1e-4 & LF     & {0.300} & 0.276 & 0.264 & 0.270 \\
1e-4 & ID mAP & 0.432 & 0.460 & 0.472 & {0.477} \\
\midrule
2e-4 & RO     & 0.550 & 0.532 & 0.526 & 0.524 \\
2e-4 & RA     & {0.298} & 0.269 & 0.259 & 0.246 \\
2e-4 & LF     & 0.280 & 0.248 & 0.249 & 0.257 \\
2e-4 & ID mAP & {0.446} & {0.473} & {0.487} & {0.494} \\
\bottomrule
\end{tabular}
\end{table}

\newpage
\paragraph{Exponential Moving Average.}
We keep a running average of the weights instead of the raw ones, which raises
OOD performance and reduces its variation across seeds
(\cref{tab:ema}, \cref{tab:ema_id}).
At every training step, instead of keeping the raw weights, maintain an average of old weights that mixes the old weights and the new weights.

$${weights}_{new} = 0.995 \times old_{average} + 0.005 \times current_{weights}$$

\begin{table}[!h]
\centering\small
\caption{Weight averaging against raw weights, DINO-DETR ResNet-50 on inpainted NuImages. Mean $\pm$ standard deviation over three seeds ($242$, $444$, $2244$) and
AP\textsubscript{50..95} in percent.}
\label{tab:ema}
\setlength{\tabcolsep}{3pt}
\begin{tabular}{l ccc ccc}
\toprule
& \multicolumn{3}{c}{\textbf{with EMA}} & \multicolumn{3}{c}{\textbf{without EMA}} \\
\cmidrule(lr){2-4}\cmidrule(lr){5-7}
\textbf{Epoch} & RO & RA & LF & RO & RA & LF \\
\midrule
1 & $49.3${\scriptsize$\pm0.7$} & $26.8${\scriptsize$\pm0.8$} & $36.1${\scriptsize$\pm0.9$}
  & $46.4${\scriptsize$\pm1.9$} & $26.3${\scriptsize$\pm1.3$} & $31.7${\scriptsize$\pm2.8$} \\
2 & $48.7${\scriptsize$\pm0.1$} & $26.0${\scriptsize$\pm0.1$} & $31.7${\scriptsize$\pm1.2$}
  & $47.2${\scriptsize$\pm2.6$} & $24.6${\scriptsize$\pm0.9$} & $27.5${\scriptsize$\pm2.1$} \\
3 & $47.6${\scriptsize$\pm0.6$} & $25.2${\scriptsize$\pm0.5$} & $28.9${\scriptsize$\pm1.4$}
  & $46.8${\scriptsize$\pm2.1$} & $24.4${\scriptsize$\pm1.1$} & $27.8${\scriptsize$\pm1.9$} \\
4 & $47.7${\scriptsize$\pm1.0$} & $24.5${\scriptsize$\pm0.4$} & $27.5${\scriptsize$\pm0.8$}
  & $46.3${\scriptsize$\pm1.1$} & $23.7${\scriptsize$\pm1.1$} & $26.0${\scriptsize$\pm1.5$} \\
\bottomrule
\end{tabular}
\end{table}

\begin{table}[!h]
\caption{ID validation mAP}
\label{tab:ema_id}
\centering
\small
\setlength{\tabcolsep}{2pt}
\begin{tabular}{c cccc}
\toprule
Epoch & s444 no EMA & s444 EMA & s2244 no EMA & s2244 EMA \\
\midrule
1 & 34.0 & 34.4 & 34.3 & 33.7 \\
2 & 38.0 & 38.6 & 37.5 & 38.8 \\
3 & 39.4 & 40.6 & 40.4 & 40.5 \\
4 & 41.1 & 41.4 & 41.1 & 41.2 \\
\bottomrule
\end{tabular}
\end{table}

\paragraph{Seed Variance.}
We repeat training with different seeds $n$ to establish how much of a difference
between configurations can be attributed to run-to-run variation
(\cref{tab:seed-variance}).

\begin{table}[h]
\centering\small
\caption{Run-to-run variation at the reported epoch. Mean $\pm$ standard
deviation over independently seeded runs, AP\textsubscript{50..95} in percent.}
\label{tab:seed-variance}
\setlength{\tabcolsep}{4pt}
\begin{tabular}{llcccc}
\toprule
\textbf{Model} & \textbf{Training data} & $n$ & RO & RA & LF \\
\midrule
DD:R50    & NuImages       & 3 & $48.7${\scriptsize$\pm0.11$} & $26.0${\scriptsize$\pm0.08$} & $31.7${\scriptsize$\pm1.15$} \\
GD:Swin-T & + Cityscapes   & 2 & $55.6${\scriptsize$\pm0.57$} & $28.2${\scriptsize$\pm1.34$} & $33.5${\scriptsize$\pm0.14$} \\
\bottomrule
\end{tabular}
\end{table}

\newpage
\paragraph{Outlier vocabulary.}
\label{full-voabulary}
The full outlier vocabulary in \cref{tab:vocabulary}: 90 object categories returned by ChatGPT
for the query 'give a list of objects that are unusual for a street scene'.
\begin{table}[h]
\centering\small
\caption{The full outlier vocabulary}
\label{tab:vocabulary}
\setlength{\tabcolsep}{6pt}
\begin{tabular}{@{}lll@{}}
\toprule
alligator & fox & Pots pans \\
antelope & gerbil & Printer \\
armadillo & giraffe & pylon \\
badger & Glasses & pylon large \\
bear & goat & pylong small \\
bison & gorilla & rabbit \\
bloated plastic bag & guinea pig & raccoon \\
Bookshelf & hamster & rat \\
buffalo & hedgehog & rhinoceros \\
cardboard & helicopter & robot \\
cat & hippopotamus & sea lion \\
cavy & horse & seal \\
Chair & kangaroo & sheep \\
chicken & koala & Shoes \\
chimpanzee & Lamps & skateboard \\
Clock & lemur & sloth \\
Coffee table & lion & sloth bear \\
crate & manatee & Sofa \\
crate small & mole & squirrel \\
crocodile & monkey & styrofoam \\
Cutting board & monster & tamarin \\
deer & mouse & tapir \\
Desk & okapi & tiger \\
Dishes & opossum & tire \\
dog & orangutan & tortoise \\
elephant & ostrich & Vacuum cleaner \\
emu & panda & walrus \\
Fan & penguin & weasel \\
ferret & platypus & wolf \\
flamingo & porcupine & zebra \\
\bottomrule
\end{tabular}
\end{table}

\section{Additional ablations}
\label{additonal}
\paragraph{Results on OoDIS.}
\label{size-ap}
In \cref{tab:size-ap} we report the benchmark results by object size, showing which size range each benchmark is dominated by.

\begin{table}[!h]
\centering
\small
\caption{OoDIS results, averaged over two seeds per model.
S, M and L are the COCO area ranges.}
\label{tab:size-ap}
\setlength{\tabcolsep}{4pt}
\begin{tabular}{l ccc ccc ccc}
\toprule
& \multicolumn{3}{c}{\textbf{RoadObstacle}}
& \multicolumn{3}{c}{\textbf{RoadAnomaly}}
& \multicolumn{3}{c}{\textbf{Fishyscapes}} \\
\cmidrule(lr){2-4}\cmidrule(lr){5-7}\cmidrule(lr){8-10}
\textbf{Method} & S & M & L & S & M & L & S & M & L \\
\midrule
\rowcolor{gray!20} SynOE-OD (DD:ResNet50)  & 38.7 & 69.6 & 76.3 &  3.7 & 13.6 & 60.9 & 18.3 & 59.3 & 52.7 \\
\rowcolor{gray!20} SynOE-OD (DD:Swin-L)    & \textbf{47.4} & \textbf{76.1} & \textbf{90.2} &  2.5 & 10.9 & 61.2 & 19.0 & 58.2 & 59.4 \\
\rowcolor{gray!20} SynOE-OD (GDINO:Swin-T) & 46.1 & 74.7 & 87.0 & \textbf{7.4} & \textbf{15.5} & \textbf{64.2} & \textbf{19.7} & \textbf{66.5} & \textbf{61.2} \\
\bottomrule
\end{tabular}
\end{table}

\paragraph{Synthetic outliers vocabulary.}
Nine out of 20 overlapping classes are essentially road debris, i.e., pylons, 
cardboard, styrofoam, tires etc., and are the objects that RoadObstacle and LostAndFound contain, and eleven are animals which RoadAnomaly contains.
The number of OOD annotations removed for the two setups is ($2{,}741$ and $2{,}772$) 
respectively, and the image count is identical.

\paragraph{Unlabeled synthetic outliers.}
Synthetic outliers can end up in the images without an annotation if the annotation step queries GDINO on the full image, where it often misses those outliers. 
This ablation shows that outliers left unlabeled conflict with the learning of the OOD class, therefore justifying our design choice in \cref{assign_label}. 
We take the same 20 overlapping classes as above and, instead of discarding them, 
leave them unannotated in the images. 
RoadAnomaly falls from AP of $26.0$\% to $20.0$\% for DINO-DETR and from $29.3$\% to $20.1$\% for GDINO \cref{tab:oe-unlabelled}.  
The reason is the mechanism described in the introduction, 
that unlabeled objects supervise against the OOD class and here measured for synthetic outliers.

\begin{table}[h]
\caption{Effect of 2{,}741 unlabeled outliers. The parentheses show the difference from the baseline.}
\label{tab:oe-unlabelled}
\centering
\small
\setlength{\tabcolsep}{2pt}
\begin{tabular}{lccc}
\toprule
\textbf{Model} & RO & RA & LF \\
\midrule
R50 & 0.483\dlt{-0.4} & 0.200\dlt{-6.0} & 0.300\dlt{-2.0} \\
GD  & 0.514\dlt{-2.9} & 0.201\dlt{-9.2} & 0.268\dlt{-2.4} \\
\bottomrule
\end{tabular}
\end{table}

\paragraph{ID/OOD trade-off.}
In \cref{tab:ID-perclass-ann} we report per-class ID performance alongside the
annotation counts before and after inpainting. \textit{Pedestrian} and
\textit{bus} keep all of their annotations and lose the least.

\begin{table}[h]
\caption{Per class ID performance in AP\textsubscript{50..95}. The
\textit{train} annotations are used for fine-tuning SynOE-OD and the \textit{original} annotations for the ID baseline.
The values in parentheses show the difference from the baseline.}
\label{tab:ID-perclass-ann}
\centering\small
\setlength{\tabcolsep}{4pt}
\begin{tabular}{lrrcc}
\toprule
\textbf{Class} & original & train
& ID baseline & SynOE-OD \\
\midrule
car        & 46{,}745  & 38{,}100 & 58.3 & 57.1\dlt{-1.2} \\
pedestrian & 32{,}210  & 32{,}210 & 42.3 & 41.4\dlt{-0.9} \\
truck      &  8{,}029  &  5{,}386 & 42.8 & 40.6\dlt{-2.2} \\
motorcycle &  3{,}406  &  2{,}638 & 47.3 & 44.9\dlt{-2.4} \\
bus        &  1{,}987  &  1{,}987 & 42.4 & 41.7\dlt{-0.7} \\
bicycle    &  2{,}285  &  1{,}447 & 35.8 & 32.3\dlt{-3.5} \\
trailer    &     636   &     429  & 16.6 & 13.4\dlt{-3.2} \\
\midrule
mAP        &     --      &    --      & 40.8 & 38.8\dlt{-2.0} \\
\bottomrule
\end{tabular}
\end{table}

\paragraph{Annotation refinement.}
We train on the unrefined inpainting boxes and without label reassignment to
isolate what each of the two refinement steps contributes
(\cref{tab:unrefined}).

\begin{table}[h]
\caption{Effect of bounding-box refinement (left) and label reassignment (right). Left: \emph{Unrefined}
refers to the original boxes and \emph{refined} is the accurate boxes returned by GDINO.
Right: \emph{replaced} is the number of ID objects of each class that was inpainted over,
\emph{reassigned} is the number for which GDINO identified the regenerated object as the
original class; the last two columns show the class AP without and with the label reassignment.}
\label{tab:unrefined}
\centering\footnotesize
\begin{subtable}[h]{0.48\linewidth}
\centering
\setlength{\tabcolsep}{3pt}
\begin{tabular}{lcccc}
\toprule
\textbf{OOD boxes} & \textbf{RO} & \textbf{RA} & \textbf{LF} & \textbf{ID} \\
\midrule
unrefined & 1.3  & 4.2  & 1.9  & 33.7 \\
refined   & 46.9 & 24.4 & 29.3 & 34.5 \\
\bottomrule
\end{tabular}
\end{subtable}
\begin{subtable}[h]{0.48\linewidth}
\centering
\setlength{\tabcolsep}{3pt}
\begin{tabular}{lrrcc}
\toprule
\textbf{Class} & \textbf{replaced} & \textbf{reassigned} & \textbf{w/o} & \textbf{with} \\
\midrule
car   & 9{,}952 & 1{,}307 & 51.0 & \textbf{54.7} \\
\bottomrule
\end{tabular}
\end{subtable}
\end{table}

\newpage
\section{Qualitative results.}
\label{qualitative}
In this section, we visualize the qualitative results. The detections in orange represent the OOD class, while the remaining predictions correspond to ID objects.

\begin{figure}[h]
    \centering
    \includegraphics[width=\linewidth]{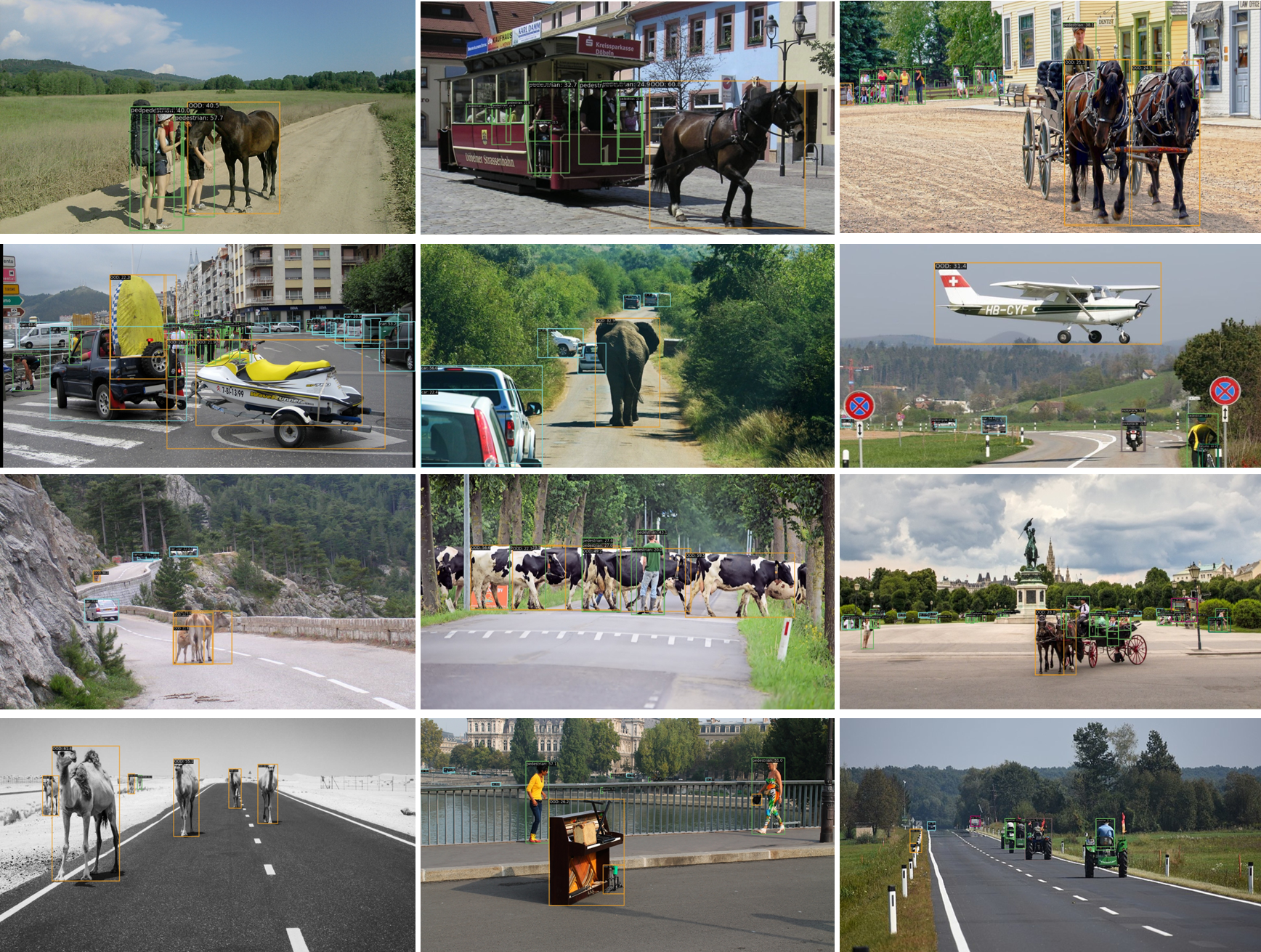}
    \caption{OOD and ID object detection results on the RoadAnomaly dataset.}
    \label{fig:placeholder}
\end{figure}

\begin{figure}[h]
    \centering
    \includegraphics[width=\linewidth]{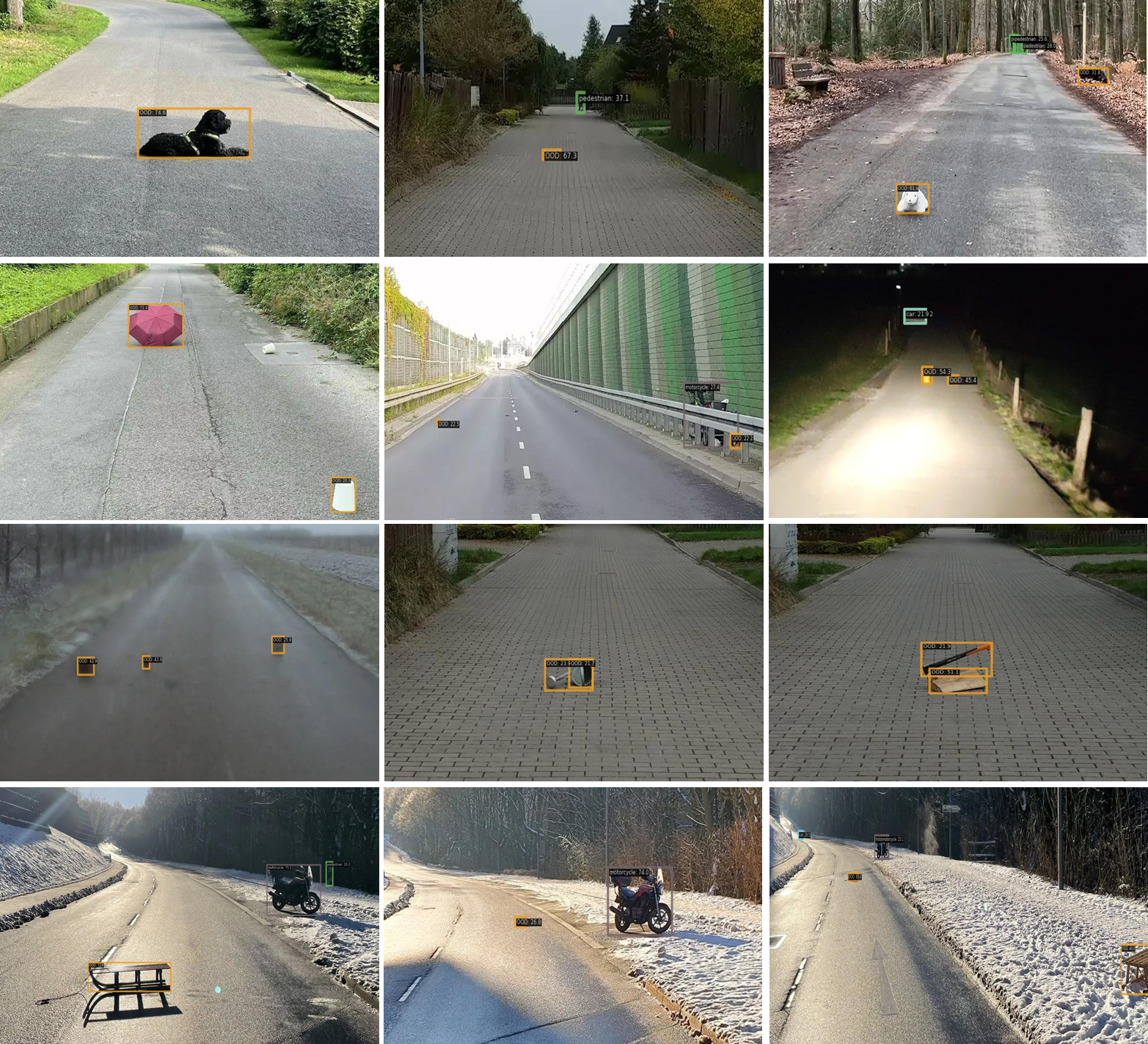}
    \caption{OOD and ID object detection results on the RoadObstacle dataset.}
    \label{fig:placeholder}
\end{figure}

\end{document}